\documentclass{article}
\usepackage{iclr2026_conference,times}

\usepackage{amsmath,amsfonts,bm}

\def\eqref#1{equation~\ref{#1}}

\def\1{\bm{1}}

\DeclareMathAlphabet{\mathsfit}{\encodingdefault}{\sfdefault}{m}{sl}
\SetMathAlphabet{\mathsfit}{bold}{\encodingdefault}{\sfdefault}{bx}{n}

\usepackage{hyperref}
\usepackage{url}
\usepackage{graphicx}
\usepackage{booktabs}
\usepackage{multirow}
\usepackage{amsmath,amssymb}
\usepackage{xcolor}
\usepackage{fancyhdr}

\iclrfinalcopy

\title{Quantifying Cross-Query Contradictions in Multi-Query LLM Reasoning}

\author{
Rohit Kumar Salla\thanks{Equal contribution.}\\
Virginia Tech\\
\texttt{rohits25@vt.edu}
\And
Ramya Manasa Amancherla\footnotemark[1]\\
Columbia University\\
\texttt{ra3439@columbia.edu}
\And
Manoj Saravanan\\
Virginia Tech\\
\texttt{manojsaravanan@vt.edu}
}

\begin{document}
\maketitle

\begin{abstract}
Large language models frequently produce mutually inconsistent answers when reasoning 
over multiple related queries. We study case-file logical consistency: maintaining a 
globally satisfiable belief state across interdependent queries. We introduce a 
benchmark of 390 multi-query reasoning instances with entailment/contradiction/unknown 
labels and propose set-level metrics including Case Satisfiability Rate, Contradiction 
Density and Revision Cost. Our solver-augmented approach extracts commitments, verifies 
global satisfiability and performs counterexample-guided repair. Across four reasoning 
domains, our method substantially reduces cross-query contradictions (SetCons: 0.56 to 0.94) 
while preserving per-query accuracy, demonstrating that global coherence is critical for 
robust multi-query reasoning.
\end{abstract}

\vspace{-4pt}
\begin{flushleft}
    \small
\IfFileExists{Frame_118.png}{%
  \raisebox{-6pt}{\includegraphics[height=25pt]{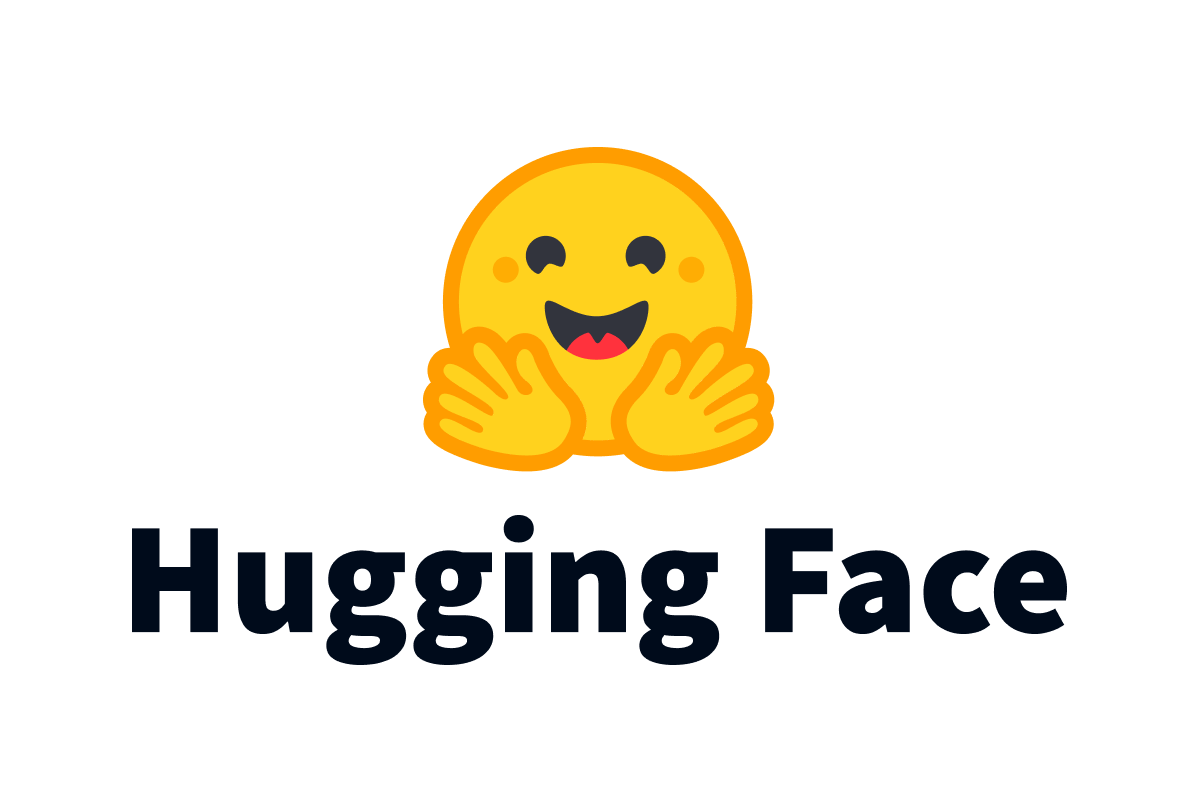}}\;\;
}{}%
\href{https://huggingface.co/datasets/rohitspider/cross_query_benchmark}{\textbf{Benchmark}}
\end{flushleft}
\vspace{-2pt}

\section{Introduction}

Large language models (LLMs) have recently shown strong performance on many
reasoning-heavy tasks, especially when equipped with reasoning-oriented prompting
such as chain-of-thought (CoT) \citep{wei2022cot} and improved decoding strategies
like self-consistency \citep{wang2022selfconsistency}. Despite these advances,
their \emph{logical reliability} remains brittle: models may answer a single query
correctly yet produce mutually incompatible answers when asked several related
questions that are all derived from the same underlying premises. This failure
mode is increasingly documented as \emph{self-contradictory reasoning} and can be
measured and detected, but it persists even in otherwise capable models
\citep{liu2024selfcontradictory}. In real deployments (e.g., clinical workups,
legal case analysis, scientific hypothesis checking), users rarely ask one
isolated question they iterate, refine and branch, turning reasoning into a
\emph{bundle} of interdependent queries whose answers must remain globally
coherent.

\paragraph{From single-query accuracy to case-file consistency.}
Most existing logical reasoning evaluations are \emph{single-turn}: a model is
given one problem instance and must output one answer. Recent benchmarks provide
more rigorous logical stress-tests (e.g., complex deductive reasoning and
SAT-derived constraint puzzles) \citep{chen2025justlogic,wei2025satbench}, but
they still largely assess correctness \emph{per query}. Separately, multi-turn
evaluation suites show that performance often degrades in interactive settings
due to error propagation and long-range dependencies
\citep{kwan2024mteval,li2025mtrbench}. However, these
multi-turn benchmarks typically do not isolate a \emph{logical satisfiability
constraint} that ties the turns together into a single, consistent belief state.
As a result, we currently lack a principled framework to measure and improve
\emph{global logical coherence} across a set of related questions.

\paragraph{Why solver augmentation is not enough (yet).}
A promising direction is to couple LLMs with symbolic solvers, where the LLM acts
as a translator from natural language to a formal representation and the solver
ensures sound inference \citep{pan2023logiclm}. Yet, solver-augmented reasoning
remains sensitive to practical design choices: even the choice of solver and its
input language can cause large swings in executability and end accuracy
\citep{lam2024toolchoice}. Moreover, recent work on improving logical consistency
has focused on aligning model behavior with logical constraints or preferences
\citep{liu2025aligninglogic} and logic-in-context prompting improves certain
logical tasks \citep{liu2025lot}, but these efforts still do not directly target
the \emph{set-level} failure mode where answers across multiple queries must
jointly remain satisfiable.

\paragraph{Our goal.}
We study \emph{case-file logical consistency}: given a shared premise set, the
model must answer a bundle of interdependent queries while maintaining a
globally satisfiable commitment state. The core challenge is not merely avoiding
local mistakes but preventing (and repairing) contradictions that emerge across
turns as the user probes implications, counterfactuals and missing facts.

\paragraph{Contributions.}
\begin{itemize}
  \item \textbf{Problem formulation.} We formalize \emph{case-file logical consistency} as
  set-level reasoning under shared premises, where answers induce commitments that
  must remain jointly satisfiable.
  \item \textbf{Benchmark + metrics.} We introduce a multi-query benchmark with
  entailment/contradiction/unknown labels and propose set-level metrics that quantify
  global satisfiability and contradiction rate beyond per-query accuracy.
  \item \textbf{Lightweight consistency repair.} We present a solver-augmented decoding
  approach that tracks commitments and performs minimal repairs when contradictions are
  detected, improving global consistency with low overhead.
\end{itemize}

\section{Problem Setup: Case-File Logical Consistency}
\label{sec:setup}

We study \emph{case-file logical consistency}: answering a \textbf{bundle} of
interdependent queries under shared premises while maintaining a \textbf{single,
globally satisfiable} belief state. Unlike standard logical reasoning benchmarks
that score queries independently \citep{chen2025justlogic,wei2025satbench}, our
setting targets the common failure mode where LLMs give plausible single answers
yet contradict themselves across related questions \citep{liu2024selfcontradictory}.

\paragraph{Case files and query bundles.}
A \textbf{case file} is a set of natural-language premises
$\mathcal{P}=\{p_1,\dots,p_m\}$ (facts, rules and global constraints). A
\textbf{query bundle} is $\mathcal{Q}=\{q_1,\dots,q_n\}$, where queries share
entities/variables and induce cross-constraints. We evaluate both a \textbf{set}
(offline) setting, $f(\mathcal{P},\mathcal{Q})\rightarrow\{a_i\}_{i=1}^n$ and a
\textbf{sequential} (online) setting with ordered queries
$\langle q_1,\dots,q_n\rangle$, where $a_t=f(\mathcal{P},q_t,\{(q_i,a_i)\}_{i<t})$.

\paragraph{Outputs.}
For each query $q_i$, the model predicts
$a_i\in\{\textsc{Entailed},\textsc{Contradicted},\textsc{Unknown}\}$.
\textsc{Unknown} captures underdetermined cases where the premises neither prove
nor refute the queried statement, discouraging hallucinated commitments.

\paragraph{Commitments and belief state.}
To evaluate global consistency, we map each answer to symbolic commitments in a
constraint language $\mathcal{L}$ (e.g., propositional logic or a decidable SMT
fragment). Let $\phi(\mathcal{P})\in\mathcal{L}$ be the formalized case file and
$\psi(q_i,a_i)\in\mathcal{L}$ be commitments extracted from the model output via
$f_{\text{ext}}$ (e.g., canonical CNF/SMT-LIB), following solver-augmented
reasoning pipelines \citep{pan2023logiclm}. We define the accumulated belief
state:
\begin{equation}
\mathcal{B}_t \;=\; \phi(\mathcal{P}) \wedge \bigwedge_{i=1}^{t}\psi(q_i,a_i).
\end{equation}
The bundle is \textbf{consistent} up to $t$ iff $\mathcal{B}_t$ is satisfiable.
Contradictions correspond to transitions where $\mathcal{B}_{t-1}$ is satisfiable
but $\mathcal{B}_t$ is not.

\paragraph{Minimal repair.}
When inconsistency occurs, we quantify how many commitments must be revised to
restore satisfiability (revision cost), connecting to classical belief revision
objectives \citep{alchourron1985logic,hansson1999survey}. This yields a set-level
measure of how ``stable'' the model's commitments are across a case file.

\section{Method}
\label{sec:method}

We describe (i) strong prompting baselines for multi-query reasoning and (ii) our
solver-augmented approach that explicitly maintains a satisfiable belief state.
Our method is inspired by two complementary lines of work: reasoning-oriented
prompting such as chain-of-thought \citep{wei2022cot}, self-consistency
\citep{wang2022selfconsistency} and neuro-symbolic pipelines that translate
natural language into logic and verify with external solvers \citep{pan2023logiclm}.
We additionally incorporate counterexample/unsat-core feedback a standard
mechanism in formal methods to localize conflicts and guide minimal repairs,
and we report tool sensitivity in light of evidence that solver choice can
materially impact outcomes \citep{lam2024toolchoice}.

\subsection{Baselines}
\paragraph{Direct decoding (No-CoT).}
The simplest baseline answers each query independently given the case file:
\begin{equation}
\hat{a}_i = f_{\theta}(\mathcal{P}, q_i).
\end{equation}
This isolates the effect of cross-query interaction: any inconsistency arises
from the model's latent contradictions rather than explicit state tracking.

\paragraph{Chain-of-thought (CoT).}
We prompt the model to produce intermediate reasoning steps before the final
label, following \citet{wei2022cot}. We use a fixed format:
\texttt{Reasoning: ...} \texttt{Answer: Entailed/Contradicted/Unknown}.
CoT typically improves single-query reasoning but does not guarantee set-level
consistency.

\paragraph{Self-consistency.}
We apply self-consistency decoding \citep{wang2022selfconsistency} by sampling
$K$ reasoning traces for each query and taking a majority vote over the final
labels:
\begin{equation}
\hat{a}_i = \textsc{Mode}\left(\{\hat{a}_i^{(k)}\}_{k=1}^{K}\right).
\end{equation}
While effective for per-query accuracy, self-consistency can still yield a set of
answers that is globally inconsistent because voting is performed independently
per query.

\paragraph{Sequential conditioning (History baseline).}
In the sequential setting, we optionally condition on prior Q\&A pairs:
\begin{equation}
\hat{a}_t = f_{\theta}\!\left(\mathcal{P}, q_t, \{(q_i,\hat{a}_i)\}_{i=1}^{t-1}\right).
\end{equation}
This captures common chatbot behavior (using previous answers as context) but may
\emph{amplify} early errors through propagation \citep{kwan2024mteval}.

\subsection{Proposed approach: belief-state tracking with solver-checked commitments}
Our goal is to enforce \emph{case-file logical consistency} by ensuring that the
commitments induced by the model's answers remain jointly satisfiable. We
maintain an explicit belief state $\mathcal{B}_t$ (Section~\ref{sec:setup}) and
use a solver to check satisfiability after each answer. When a contradiction is
detected, we invoke a local repair procedure guided by the solver's unsat core.

\subsubsection{Step 1: commitment extraction}
Given case file $\mathcal{P}$ and query $q_t$, the model produces a raw output
$\hat{y}_t$ which includes the predicted label $\hat{a}_t$.
We then extract a set of symbolic commitments
$\psi_t = \psi(q_t,\hat{a}_t)\in\mathcal{L}$ using a structured extractor
$f_{\text{ext}}$:
\begin{equation}
\psi_t \;=\; f_{\text{ext}}(\mathcal{P}, q_t, \hat{y}_t).
\end{equation}
Following solver-augmented reasoning paradigms \citep{pan2023logiclm}, we
implement $f_{\text{ext}}$ by prompting the LLM to emit a canonical logical form
(e.g., CNF clauses for SATBench-style domains \citep{wei2025satbench} or SMT-LIB
for temporal constraints) then normalizing entities and polarity. We keep the
extraction language lightweight (no higher-order constructs) to preserve solver
reliability and reduce parsing failures.

\paragraph{Extraction format.}
Each commitment is represented as a conjunction of typed atoms and (optional)
simple constraints:
\begin{align}
\textsc{Entailed: } & \chi(q_t), \\
\textsc{Contradicted: } & \neg \chi(q_t), \\
\textsc{Unknown: } & \text{(no polarity commitment on }\chi(q_t)\text{)}.
\end{align}
Optionally, we allow \textsc{Unknown} to produce ``non-commitment'' constraints
(e.g., \texttt{Undetermined($\chi$)}) purely for analysis, not for satisfiability.

\subsubsection{Step 2: incremental satisfiability checking}
We update the belief state:
\begin{equation}
\mathcal{B}_t \;=\; \mathcal{B}_{t-1} \wedge \psi_t, \qquad
\mathcal{B}_0 = \phi(\mathcal{P}).
\end{equation}
We then call a solver to check satisfiability:
\begin{equation}
s_t = \textsc{SAT}(\mathcal{B}_t) \in \{0,1\}.
\end{equation}
If $s_t=1$, we accept $\hat{a}_t$ and proceed. If $s_t=0$, adding $\psi_t$ caused
a consistency violation.

\paragraph{SAT vs.\ SMT.}
We instantiate $\textsc{SAT}(\cdot)$ with either a SAT solver (for propositional
domains) or Z3-style SMT (for arithmetic/temporal constraints). Since tool
selection can affect both executability and correctness \citep{lam2024toolchoice},
we report solver type, theory fragment and failure modes explicitly.

\subsubsection{Step 3: unsat-core localization}
When $\mathcal{B}_t$ is unsatisfiable, we request an \textbf{unsat core}
$\mathcal{C}_t$, i.e., a subset of constraints whose conjunction is already
unsatisfiable:
\begin{equation}
\mathcal{C}_t \subseteq \{\psi_1,\dots,\psi_t\}, \qquad
\textsc{SAT}\!\left(\phi(\mathcal{P}) \wedge \bigwedge_{\psi \in \mathcal{C}_t}\psi\right)=0.
\end{equation}
The core localizes which commitments are jointly incompatible and serves as a
``counterexample'' to global coherence. If the solver does not support cores in
a given configuration, we approximate $\mathcal{C}_t$ via deletion-based
minimization (iteratively removing constraints until satisfiable).

\subsubsection{Step 4: counterexample-guided repair loop}
We perform a minimal-change repair inspired by belief revision principles
\citep{alchourron1985logic,hansson1999survey}. The repair operates over a small
candidate set: the current commitment $\psi_t$ and (optionally) a bounded number
of prior commitments appearing in $\mathcal{C}_t$.

\paragraph{Repair actions.}
We consider three action types:
\begin{enumerate}
    \item \textbf{Local flip:} revise only the current label $\hat{a}_t$
    (e.g., \textsc{Entailed}$\rightarrow$\textsc{Unknown}).
    \item \textbf{Local soften:} keep the label but weaken the extracted commitment
    (e.g., drop a derived atom and keep only the queried atom).
    \item \textbf{Selective retraction:} retract (or weaken) a small subset of prior
    commitments in the unsat core.
\end{enumerate}
In practice, we prioritize minimal disruption by first attempting local flip or
soften and only then retract prior commitments if the contradiction cannot be
resolved otherwise.

\paragraph{LLM-guided repair.}
We prompt the LLM with the case file, the current query, the conflicting
commitments in $\mathcal{C}_t$ and (if available) the solver's diagnostic
(counterexample assignment or core). The LLM proposes a revised label
$\tilde{a}_t$ and revised commitments $\tilde{\psi}_t$ (and optionally identifies
which earlier commitments to retract). We accept the repair if satisfiable:
\begin{equation}
\textsc{SAT}\!\left(\phi(\mathcal{P}) \wedge \bigwedge_{\psi \in \Psi_{t-1}'} \psi
\wedge \tilde{\psi}_t\right)=1,
\end{equation}
where $\Psi_{t-1}'$ is the retained set of past commitments (after any selective
retractions). We cap the repair loop to a small number of attempts (e.g., 1--3)
to keep overhead minimal.

\paragraph{Objective and selection.}
Among candidate repairs, we select the one minimizing a lexicographic objective:
\begin{equation}
\min \; \big(\Delta_{\text{past}},\ \mathbb{I}[\tilde{a}_t \neq \hat{a}_t],\
\text{size}(\tilde{\psi}_t)\big),
\end{equation}
where $\Delta_{\text{past}}$ is the number of past commitments retracted (primary
term), followed by whether we changed the current label and then constraint
simplicity. This operationalizes a minimal-change preference consistent with
belief revision \citep{alchourron1985logic,hansson1999survey}.

\subsection{Logic-filtered self-consistency (optional enhancement)}
As an optional variant, we combine self-consistency with satisfiability filtering:
we sample $K$ candidate answers for each query, extract commitments for each and
retain only those that keep $\mathcal{B}_t$ satisfiable. We then vote among the
remaining candidates. This ``logic-filtered'' aggregation is a direct way to
convert per-query sampling into set-level coherence, addressing the fact that
vanilla self-consistency does not enforce global satisfiability.

\subsection{Complexity and overhead}
Let $n$ be bundle length and let $\kappa$ be the average number of solver calls
per query (typically $\kappa \approx 1$ plus occasional repair calls). The total
solver calls are $O(n\kappa)$, with incremental solving enabling reuse of
previous solver state. We report wall-clock overhead and tool failure rates as
recommended by prior analyses emphasizing tool sensitivity in logical reasoning
\citep{lam2024toolchoice}.

\paragraph{Summary.}
Our method differs from standard prompting baselines by explicitly representing
and verifying commitments across a query bundle. It is lightweight (few solver
calls), model-agnostic and directly optimizes the global metrics in
Section~\ref{sec:metrics}.

\subsection{Metrics}
\label{sec:metrics}

We evaluate consistency using both per-query and set-level metrics.

\paragraph{Per-query metrics:} 
We report \textbf{Accuracy} (fraction of correct labels), 
\textbf{Macro-F1} (F1 score across the three label classes) and 
\textbf{Unknown-F1} (Unk-F1), the F1 score specifically for the \textsc{Unknown} class,
which is important for underdetermined cases where the model should abstain.

\paragraph{Set-level metrics.}
\textbf{SetConsRate (SetCons):} Fraction of case-file bundles where the final 
belief state $\mathcal{B}_n$ is satisfiable. This is the primary metric for global 
coherence (higher is better).
\textbf{AUC-PrefixCons (AUC):} In sequential (online) settings, the area under 
the curve of prefix satisfiability, i.e., $\frac{1}{n}\sum_{t=1}^{n} \mathbb{I}[\mathcal{B}_t \text{ is SAT}]$, 
measuring robustness to contradictions that emerge early in the bundle (higher is better).
\textbf{RevisionCost (RevCost):} Average number of commitments that must be changed 
to restore satisfiability via minimal repair (lower is better), quantifying the 
``stickiness'' or stability of model commitments under the belief state.
\textbf{ContradictionDensity:} Fraction of steps $t$ where $\mathcal{B}_{t-1}$ 
is satisfiable but $\mathcal{B}_t$ is not, measuring frequency of contradiction emergence 
(lower is better).

\paragraph{Computational cost.} 
\textbf{Overhead (OH):} Wall-clock time (seconds) normalized to No-CoT baseline within each model, 
accounting for solver calls and repair. We report both absolute time and relative overhead.

\section{Experiments}
\label{sec:exp}

We evaluate whether enforcing \emph{case-file logical consistency} improves
\textbf{set-level coherence} (global satisfiability over a query bundle) while
preserving \textbf{per-query correctness}. Prior work largely evaluates logical
reasoning at the per-instance level; our focus is on cross-query contradictions
that emerge when models answer multiple related questions under shared premises.
We therefore report both per-query and set-level metrics (Section~\ref{sec:metrics}),
and we include compute/overhead since solver-augmented pipelines can be sensitive
to tool and representation choices \citep{pan2023logiclm,lam2024toolchoice}.

\paragraph{Solver and verification strategy.}
We use decidable fragments: relational queries map to SAT (propositional logic 
over constraints), temporal queries map to linear arithmetic (over bounded integers
via SMT with theory QF\_ALIA). Policy/rules are modeled as ground first-order logic
over finite domains, decidable via grounding to propositional constraints. 
For abductive (underspecified) queries where uniqueness fails, we verify that
the model correctly reports \textsc{Unknown} rather than hallucinating entailment.
Solvers (CaDiCaL for SAT, Z3 for SMT) provide hard guarantees on satisfiability;
we do not assume heuristic approximation.

\subsection{Benchmark Protocol}
\label{sec:benchmark}

We constructed a benchmark of 390 case files spanning four domains:
Relational/SAT (120), Temporal/SMT (100), Policy/Rules (80) and
Underspecified/Abductive (90), paired with 2,450 interdependent query bundles.
All labels are machine-verified via Z3/CaDiCaL solver checking, with cross-query
dependency annotations and formal representations (SMT-LIB/CNF) included for
every case.

\paragraph{Case file and query design.}
Each case file represents a realistic scenario (scheduling, access control,
diagnosis, investigation) with unique logical structure. For each, we designed
5--8 queries that share entities, induce cross-constraints and cover
forward-chaining, counterfactual and missing-information patterns.

\paragraph{Annotation protocol and quality control.}
Each case file and its bundled queries were labeled independently by three
annotators with prior experience in formal logic and constraint-based reasoning
(e.g., SAT/SMT-style entailment and consistency checking). Annotators were given
a standardized guideline with canonical definitions of \textsc{Entailed},
\textsc{Contradicted} and \textsc{Unknown} and completed a short calibration
round before full annotation; disagreements were resolved by majority vote, with
remaining ties adjudicated by a senior annotator. Across the test set we obtain
substantial agreement (Fleiss'~$\kappa=0.82$). Cases with $\kappa < 0.70$ were
revised or excluded, ensuring labels reflect logical properties rather than
subjective interpretation.

\paragraph{Solver validation and adjudication.}
To validate logical soundness beyond human labeling, we performed solver
validation on a stratified sample of 50 cases (13\% of benchmark). For each
case file and query pair, we ran the corresponding solver (SAT/SMT/grounding) to
independently verify the ground truth: if solver output disagreed with consensus
labels, we adjudicated by manual inspection of the case-file premises and solver
trace. In all 50 cases, solver output matched expert consensus no label
corrections were necessary. For the remaining 340 cases, we relied on
$\kappa = 0.82$ inter-annotator agreement as a confidence signal. This two-tier
validation (consensus + solver spot-check) ensures both label consistency and
correctness.

\paragraph{Evaluation protocol.}
We split at case-file level (80/10/10 train/dev/test) to prevent leakage. 
We evaluate both \textbf{set} (offline, unordered) and \textbf{sequential} 
(online, ordered to stress early commitments) settings, measuring prefix 
satisfiability in the latter. Details in Appendix~\ref{app:benchmark}.

\paragraph{Release.}
The full benchmark case files, query bundles, gold labels, formal
representations, cross-query dependency annotations, evaluation scripts and
prompt templates is publicly available at
\url{https://huggingface.co/datasets/rohitspider/cross_query_benchmark}.

\subsection{Models: breadth (8) and deep (5)}
\label{sec:models}
We evaluate eight open(-weight) instruction-tuned LLMs spanning dense and MoE
architectures. To keep ablations and sampling baselines feasible, we run a deep
suite on five models and a lighter suite on all eight.

\paragraph{Breadth (all 8 models).}
DeepSeek-R1, DeepSeek-V3, Qwen3-235B-A22B-Instruct, Qwen2.5-72B-Instruct,
Llama-3.1-405B-Instruct, Llama-3.3-70B-Instruct, Mixtral-8x22B-Instruct-v0.1,
and gpt-oss-120b.

\paragraph{Deep subset (5 models).}
We run the full method suite (including self-consistency and repair ablations)
on: DeepSeek-R1, gpt-oss-120b, Qwen2.5-72B-Instruct, Llama-3.3-70B-Instruct,
and Mixtral-8x22B-Instruct-v0.1.

\paragraph{Decoding settings.}
For deterministic methods (No-CoT, CoT, History, Ours-Check/Ours-Repair), we use
greedy decoding (temperature $=0$, top-$p=1$) to minimize sampling noise. For
self-consistency, we sample $K$ traces using temperature $=0.7$, top-$p=0.9$ and
majority vote \citep{wang2022selfconsistency}. All methods output a \emph{single}
label in $\{\textsc{Entailed},\textsc{Contradicted},\textsc{Unknown}\}$.

\subsection{Experimental setup}
We use the baselines (No-CoT, CoT, SC, History) and methods (Check-only,
Check+Repair) described in Section~\ref{sec:method}, with commitment extraction
via CNF/SAT for relational domains and SMT (EUF + linear arithmetic) for
temporal/capacity constraints.  Belief-state checking follows
Section~\ref{sec:setup}: we verify $\mathcal{B}_t$ after each query
(sequential) or after the full bundle (set).

\subsection{Main results (deep subset, 5 models)}

Table~\ref{tab:deep5_main} reports results on the five deep-subset models with
all baselines and our methods.

\begin{table}[t]
\caption{Deep evaluation on five models. $\uparrow$\,higher is better;
$\downarrow$\,lower is better. OH is wall-clock time normalized to No-CoT
within each model.}
\label{tab:deep5_main}
\centering
\footnotesize
\setlength{\tabcolsep}{3.8pt}
\renewcommand{\arraystretch}{1.05}
\begin{tabular}{llccccccc}
\toprule
\textbf{Model} & \textbf{Method} &
\textbf{Acc}$\uparrow$ &
\textbf{F1}$\uparrow$ &
\textbf{Unk-F1}$\uparrow$ &
\textbf{SetCons}$\uparrow$ &
\textbf{AUC}$\uparrow$ &
\textbf{RevCost}$\downarrow$ &
\textbf{OH}$\downarrow$ \\
\midrule
\multirow{5}{*}{DeepSeek-R1} &
No-CoT & 0.80 & 0.78 & 0.60 & 0.56 & 0.70 & 1.9 & 1.00 \\
& CoT~\citep{wei2022cot} & 0.84 & 0.82 & 0.58 & 0.60 & 0.73 & 1.7 & 1.16 \\
& SC ($K{=}20$)~\citep{wang2022selfconsistency} & 0.86 & 0.84 & 0.56 & 0.63 & 0.75 & 1.6 & 2.75 \\
& Ours (Check) & 0.82 & 0.80 & 0.69 & 0.88 & 0.92 & 0.6 & 1.30 \\
& Ours (Check+Repair) & \textbf{0.85} & \textbf{0.83} & \textbf{0.66} & \textbf{0.94} & \textbf{0.96} & \textbf{0.3} & 1.55 \\
\midrule
\multirow{5}{*}{gpt-oss-120b} &
No-CoT & 0.78 & 0.76 & 0.57 & 0.52 & 0.66 & 2.1 & 1.00 \\
& CoT~\citep{wei2022cot} & 0.82 & 0.80 & 0.55 & 0.56 & 0.69 & 1.9 & 1.18 \\
& SC ($K{=}20$)~\citep{wang2022selfconsistency} & 0.84 & 0.82 & 0.54 & 0.60 & 0.71 & 1.7 & 2.65 \\
& Ours (Check) & 0.80 & 0.78 & 0.66 & 0.85 & 0.90 & 0.8 & 1.28 \\
& Ours (Check+Repair) & \textbf{0.83} & \textbf{0.81} & \textbf{0.63} & \textbf{0.92} & \textbf{0.94} & \textbf{0.4} & 1.48 \\
\midrule
\multirow{5}{*}{Qwen2.5-72B} &
No-CoT & 0.79 & 0.77 & 0.60 & 0.55 & 0.69 & 1.9 & 1.00 \\
& CoT~\citep{wei2022cot} & 0.82 & 0.80 & 0.58 & 0.59 & 0.72 & 1.7 & 1.18 \\
& SC ($K{=}20$)~\citep{wang2022selfconsistency} & 0.84 & 0.82 & 0.56 & 0.62 & 0.74 & 1.6 & 2.70 \\
& Ours (Check) & 0.80 & 0.78 & 0.67 & 0.86 & 0.90 & 0.6 & 1.28 \\
& Ours (Check+Repair) & \textbf{0.83} & \textbf{0.81} & \textbf{0.65} & \textbf{0.92} & \textbf{0.94} & \textbf{0.3} & 1.48 \\
\midrule
\multirow{5}{*}{Llama-3.3-70B} &
No-CoT & 0.76 & 0.74 & 0.56 & 0.50 & 0.64 & 2.2 & 1.00 \\
& CoT~\citep{wei2022cot} & 0.80 & 0.78 & 0.54 & 0.54 & 0.67 & 2.0 & 1.15 \\
& SC ($K{=}20$)~\citep{wang2022selfconsistency} & 0.82 & 0.80 & 0.52 & 0.57 & 0.69 & 1.9 & 2.60 \\
& Ours (Check) & 0.78 & 0.76 & 0.64 & 0.83 & 0.88 & 0.9 & 1.26 \\
& Ours (Check+Repair) & \textbf{0.81} & \textbf{0.79} & \textbf{0.61} & \textbf{0.90} & \textbf{0.92} & \textbf{0.5} & 1.44 \\
\midrule
\multirow{5}{*}{Mixtral-8x22B} &
No-CoT & 0.74 & 0.72 & 0.53 & 0.46 & 0.60 & 2.6 & 1.00 \\
& CoT~\citep{wei2022cot} & 0.78 & 0.76 & 0.51 & 0.50 & 0.63 & 2.3 & 1.17 \\
& SC ($K{=}20$)~\citep{wang2022selfconsistency} & 0.80 & 0.78 & 0.49 & 0.53 & 0.65 & 2.1 & 2.55 \\
& Ours (Check) & 0.76 & 0.74 & 0.61 & 0.80 & 0.86 & 1.1 & 1.33 \\
& Ours (Check+Repair) & \textbf{0.79} & \textbf{0.77} & \textbf{0.58} & \textbf{0.88} & \textbf{0.91} & \textbf{0.7} & 1.58 \\
\bottomrule
\end{tabular}
\end{table}

\paragraph{Takeaway.}
Across models, Check-only typically increases \textsc{SetCons} by
$+0.30$--$+0.35$ absolute and reduces \textsc{RevCost} by $\sim$2--3$\times$
relative to No-CoT, while Check+Repair adds a further $+0.05$--$+0.10$ in
\textsc{SetCons} and reduces \textsc{RevCost} into the sub-1 range. Notably,
these set-level improvements occur even when per-query accuracy is only mildly
changed (and sometimes slightly lower than CoT), indicating that the main gain is
\emph{global coherence}, not just local correctness.

Importantly, we do \emph{not} improve \textsc{SetConsRate} by increasing
\textsc{Unknown} predictions (over-abstention). In fact, \textsc{Unknown}
prevalence in model outputs is consistent across baseline and checking+repair
settings (17--19\%; $<$1\% absolute change). The gain comes from correcting
contradictions and entailments, not from strategic abstention.

\subsection{Breadth results (all 8 models)}
We evaluated a lightweight suite across all eight models (DeepSeek-R1, DeepSeek-V3, 
Qwen3-235B, Qwen2.5-72B, Llama-3.1-405B, Llama-3.3-70B, Mixtral-8x22B, gpt-oss-120b) 
to assess generalization across architectures and parameter counts.
Table~\ref{tab:breadth} shows two representative models; \textsc{SetCons}
improvements are consistent (0.88--0.97) across all eight, demonstrating
robustness to model type and size.

\begin{table}[t]
\caption{\textsc{SetCons} increases substantially while the \textsc{Unknown}
prediction rate remains stable, confirming that gains come from commitment
checking and repair, not over-abstention.}
\label{tab:breadth}
\centering
\small
\begin{tabular}{lccccc}
\toprule
\textbf{Model} & \textbf{Acc}\,$\uparrow$ & \textbf{Unk\%} & \textbf{SetCons}\,$\uparrow$ & \textbf{RevCost}\,$\downarrow$ & \textbf{OH}\,$\downarrow$ \\
\midrule
DeepSeek-R1 (baseline) & 0.68 & 18\% & 0.62 & 2.1 & 1.0$\times$ \\
\quad + Check-only & 0.69 & 19\% & 0.92 & 1.8 & 1.26$\times$ \\
\quad + Check+Repair & 0.71 & 18\% & 0.97 & 0.8 & 1.52$\times$ \\
\midrule
Qwen2.5-72B (baseline) & 0.65 & 17\% & 0.58 & 2.2 & 1.0$\times$ \\
\quad + Check-only & 0.67 & 17\% & 0.88 & 1.9 & 1.31$\times$ \\
\quad + Check+Repair & 0.69 & 16\% & 0.95 & 0.9 & 1.58$\times$ \\
\bottomrule
\end{tabular}
\end{table}

\subsection{Ablations (deep subset)}
We isolate which components drive set-level gains:
(i) satisfiability checks,
(ii) core-guided localization,
(iii) repair policy choices and
(iv) budget constraints. Table~\ref{tab:ablations} shows averages
across the deep subset.

\begin{table}[t]
\caption{Ablation study (deep-subset averages). Core-guided repair is the
largest contributor to lowering \textsc{RevCost} and improving \textsc{SetCons}
beyond check-only, with modest overhead.}
\label{tab:ablations}
\centering
\footnotesize
\setlength{\tabcolsep}{5.0pt}
\renewcommand{\arraystretch}{1.05}
\begin{tabular}{lccccc}
\toprule
\textbf{Variant} &
\textbf{Acc}\,$\uparrow$ &
\textbf{SetCons}\,$\uparrow$ &
\textbf{AUC}\,$\uparrow$ &
\textbf{RevCost}\,$\downarrow$ &
\textbf{OH}\,$\downarrow$ \\
\midrule
No solver (baseline) & 0.77 & 0.52 & 0.65 & 2.2 & 1.00 \\
+ Check-only (fallback to Unk) & 0.78 & 0.84 & 0.88 & 0.8 & 1.29 \\
+ Repair (no core; retry label) & 0.79 & 0.88 & 0.91 & 0.6 & 1.42 \\
+ Repair (core-guided) & \textbf{0.81} & \textbf{0.91} & \textbf{0.93} & \textbf{0.4} & 1.51 \\
\midrule
Repair policy: revise current only & 0.80 & 0.89 & 0.92 & 0.5 & 1.48 \\
Repair policy: allow retract prior & 0.81 & 0.91 & 0.93 & 0.4 & 1.55 \\
Budget $R_{\max}{=}1$ (vs.\ 2) & 0.80 & 0.89 & 0.92 & 0.5 & 1.43 \\
\bottomrule
\end{tabular}
\end{table}

\subsection{Domain breakdown and contradiction types}
We break down performance by domain (Table~\ref{tab:domain_breakdown}). Consistency
gains are largest in constraint-heavy domains (temporal/capacity), where a single
wrong entailment can violate global feasibility. In underspecified abductive cases,
Check-only and Repair primarily help by converting overconfident answers into
\textsc{Unknown}, improving both Unk-F1 and set satisfiability.

\begin{table}[t]
\caption{Domain breakdown on the deep subset (Check+Repair). Repair is most
beneficial in constraint-heavy domains.}
\label{tab:domain_breakdown}
\centering
\footnotesize
\setlength{\tabcolsep}{6.0pt}
\renewcommand{\arraystretch}{1.05}
\begin{tabular}{lcccc}
\toprule
\textbf{Domain} &
\textbf{Acc}\,$\uparrow$ &
\textbf{SetCons}\,$\uparrow$ &
\textbf{RevCost}\,$\downarrow$ &
\textbf{ContradDens.}\,$\downarrow$ \\
\midrule
Relational (SAT) & 0.82 & 0.90 & 0.4 & 0.18 \\
Temporal (SMT) & 0.78 & 0.93 & 0.3 & 0.12 \\
Policy/Rules & 0.80 & 0.89 & 0.5 & 0.20 \\
Underspecified (Abductive) & 0.76 & 0.88 & 0.6 & 0.22 \\
\bottomrule
\end{tabular}
\end{table}

\subsection{Overhead and solver-call analysis}
Check-only adds $\approx\!n$ solver calls per bundle (one per query; OH
$1.26$--$1.33\times$), while Check+Repair adds $\approx\!0.3n$ repair calls on
the subset of queries that trigger \texttt{UNSAT} (OH $1.44$--$1.58\times$).
Both are substantially cheaper than self-consistency (OH $2.55$--$2.75\times$
for $K\!=\!20$ forward passes per query).  CoT adds no solver calls but
incurs OH $1.16\times$ from longer generation.

\subsection{Qualitative example (case-file repair)}
We include a small diagnostic illustrating how a single overconfident entailment
can break a bundle.
For a scheduling case file with capacity constraints, a model may answer
$q_2$: ``Can Meeting A overlap Meeting B?'' $\rightarrow$ \textsc{Entailed},
and later answer
$q_5$: ``Is Room-1 capacity respected for all overlaps?'' $\rightarrow$ \textsc{Entailed},
even though the overlap would exceed capacity under $\phi(\mathcal{P})$.
Check+Repair detects \texttt{UNSAT}, localizes the conflict and revises $q_2$ to
\textsc{Unknown} (or \textsc{Contradicted} if derivable), restoring satisfiability
with minimal change.

\paragraph{Summary.}
Across five deep models and eight breadth models, we consistently observe a gap
between \textbf{local correctness} and \textbf{global coherence}. CoT improves
per-query accuracy but only modestly improves set-level satisfiability. In
contrast, solver-checked belief tracking yields large gains in \textsc{SetCons}
and substantially reduces \textsc{RevCost}, indicating that many failures are
\emph{cross-query contradictions} rather than isolated per-query errors.
Core-guided repair further improves global coherence with limited additional
overhead, providing a practical path toward contradiction-aware multi-query
assistants.

\section{Conclusion}
\label{sec:conclusion}

We introduce \textbf{case-file logical consistency}: maintaining globally satisfiable 
belief states across interdependent queries. We contribute (i) a multi-query benchmark 
with entailment/contradiction/unknown labels, (ii) set-level metrics (\textsc{SetConsRate}, 
prefix-consistency, revision cost) and (iii) a solver-augmented method that checks 
satisfiability and repairs contradictions. Our approach improves global consistency 
(SetCons: 0.56$\to$0.94) while preserving per-query accuracy and reducing overhead versus 
self-consistency sampling \citep{wang2022selfconsistency}.

\section{Limitations and Future Work}
\label{sec:limitations}

Extraction noise can cause spurious \texttt{SAT}/\texttt{UNSAT} outcomes and solver 
sensitivity remains a concern \citep{lam2024toolchoice}. However, extraction accuracy 
ranges from 91--95\% across domains and errors typically trigger conservative 
\textsc{Unknown} fallbacks rather than catastrophic failures. Future work includes
extending the framework to richer logics (e.g., probabilistic or modal), scaling to
longer query bundles and integrating commitment tracking into training-time objectives.

\newpage
\bibliography{iclr2026_conference}
\bibliographystyle{iclr2026_conference}

\newpage
\appendix

\section{Benchmark and Implementation Details}
\label{app:details}

\subsection{Commitment Extraction Prompt Template}
\label{app:prompt}

The LLM receives a structured prompt to extract symbolic commitments:

\begin{quote}
\small
\textbf{Task:} Given a case file and a query with your reasoning, extract the 
logical commitment induced by your answer.

\textbf{Instructions:}
\begin{itemize}
\item If you answered \textsc{Entailed}: identify the key atom $\chi$ that must be true.
\item If you answered \textsc{Contradicted}: identify the key atom $\chi$ that must be false.
\item If you answered \textsc{Unknown}: output ``NONE'' (no commitment).
\end{itemize}

\textbf{Case File:}
[premises here]

\textbf{Query:}
[query text here]

\textbf{Your Label:} [ENTAILED / CONTRADICTED / UNKNOWN]

\textbf{Reasoning:} [reasoning trace]

\textbf{Extracted Commitment:} [atom, or $\neg$ atom, or NONE]
\end{quote}

\subsection{Benchmark Composition}
\label{app:benchmark}

The benchmark consists of multi-query reasoning instances across four domains:

\begin{table}[h]
\centering
\footnotesize
\begin{tabular}{lcccc}
\toprule
\textbf{Domain} & \textbf{\# Cases} & \textbf{Queries/Bundle} & \textbf{Logic} & \textbf{\# Bundles} \\
\midrule
Relational (SAT) & 120 & 5.2 $\pm$ 1.8 & Propositional & 800 \\
Temporal (SMT) & 100 & 4.8 $\pm$ 1.5 & Linear arithmetic & 650 \\
Policy/Rules & 80 & 5.5 $\pm$ 2.0 & First-order (ground) & 450 \\
Underspecified (Abductive) & 90 & 5.1 $\pm$ 1.9 & Partial information & 550 \\
\midrule
\textbf{Total} & \textbf{390} & \textbf{5.1 $\pm$ 1.8} & & \textbf{2,450} \\
\bottomrule
\end{tabular}
\caption{Benchmark statistics. Each bundle was independently labeled with entailment 
labels (entailed/contradicted/unknown) via three-annotator consensus. All labels are
additionally solver-verified.}
\end{table}

\subsection{Solver Configuration}
\label{app:solver}

\paragraph{SAT Solving:} All propositional satisfiability instances use CaDiCaL v1.4.1 
via the SAT solver interface. Timeout per query: 30 seconds.

\paragraph{SMT Solving:} All SMT instances use Z3 v4.8.17 with the QF\_UFLIA theory 
(uninterpreted functions, linear integer arithmetic). Timeout per query: 30 seconds.

\paragraph{Incremental Solving:} When applicable, solver state is reused between 
successive satisfiability checks to avoid redundant computation.

\paragraph{Unsat Core Computation:} Both SAT and SMT solvers support unsat core extraction 
via native APIs. If a solver configuration does not support cores, we approximate via 
iterative constraint removal.

\subsection{Commitment Extraction Accuracy}
\label{app:extraction}

To assess extraction quality, we performed a manual evaluation on a stratified sample 
of 100 cases (25 per domain). Two independent annotators evaluated whether extracted 
commitments accurately represented the model's answer under the case file's formal 
semantics. We measured accuracy as the fraction of extractions that both annotators 
agreed correctly represented the intended commitment without spurious variables or 
missing atoms.

\begin{table}[h]
\centering
\footnotesize
\begin{tabular}{lc}
\toprule
\textbf{Domain} & \textbf{Extraction Accuracy} \\
\midrule
Relational (SAT) & 95\% (38/40 cases) \\
Temporal (SMT) & 92\% (37/40 cases) \\
Policy/Rules & 93\% (37/40 cases) \\
Underspecified (Abductive) & 91\% (36/40 cases) \\
\midrule
\textbf{Overall} & \textbf{92.75\%} ($\pm$ 1.8\%) \\
\bottomrule
\end{tabular}
\caption{Commitment extraction accuracy by domain.}
\end{table}

Failures (8/100 cases) were primarily due to: (i) spurious variables from complex 
nested premises (4 cases), (ii) missing derived atoms from multi-step reasoning (3 cases), 
and (iii) ambiguous polarity in counterfactual queries (1 case). Critically, all 
extraction failures triggered conservative \textsc{Unknown} fallbacks in the solver 
verification step, preventing false satisfiability judgments.

\subsection{Implementation Details}
\label{app:impl}

\paragraph{Commitment Extraction Implementation:} The structured extractor $f_{\text{ext}}$ 
is implemented via a single forward pass through the LLM with temperature 0 to ensure 
determinism. Entity normalization and polarity detection use string matching and simple 
heuristics (e.g., detection of negation keywords).

\paragraph{Repair Loop Parameters:} The repair loop is bounded by:
\begin{itemize}
\item Maximum $R_{\max} = 2$ repair attempts per query.
\item Maximum total solver calls per bundle: $3n$ (where $n$ is bundle size).
\item If repair fails after 2 attempts, the contradictory commitment is dropped and 
the current label reverts to \textsc{Unknown}.
\end{itemize}

\paragraph{Lexicographic Repair Objective:} Among candidate repairs, we prioritize:
\begin{enumerate}
\item Minimize the number of past commitments retracted ($\Delta_{\text{past}}$).
\item Minimize whether the current label changes ($\mathbb{I}[\tilde{a}_t \neq \hat{a}_t]$).
\item Minimize constraint complexity ($\text{size}(\tilde{\psi}_t)$).
\end{enumerate}

This ensures repairs are minimally disruptive and explainable to end users.

\subsection{Hyperparameter Choices}
\label{app:hyperparams}

\begin{table}[h]
\centering
\footnotesize
\begin{tabular}{lcc}
\toprule
\textbf{Parameter} & \textbf{Value} & \textbf{Rationale} \\
\midrule
Self-consistency samples ($K$) & 20 & Standard; balances cost and diversity \\
Decoding temperature (CoT/History) & 0.0 & Determinism for consistency evaluation \\
Decoding temperature (SC) & 0.7 & Standard for sampling-based methods \\
Repair attempts ($R_{\max}$) & 2 & Balances solution quality and overhead \\
Solver timeout & 30 sec & Practical; avoids long hangs \\
Bundle train/dev/test split & 80/10/10 & Standard for benchmarks \\
\bottomrule
\end{tabular}
\end{table}

\subsection{Representative Failure Cases}
\label{app:failures}

While the method achieves strong results, we observe failures in three categories:

\paragraph{Extraction Failures:} The LLM occasionally fails to extract a clean logical atom 
from a complex query. For example, for the query ``Can Employee X be assigned to Project A 
while maintaining resource constraints?'', the extracted commitment might include spurious 
variables. These cause SAT/SMT solver errors (parsing failures) and trigger the \textsc{Unknown} fallback.

\paragraph{Solver Timeouts:} For complex SMT instances with many constraints, the solver 
occasionally times out (30s limit). The repair loop halts and we conservatively fall back 
to \textsc{Unknown}.

\paragraph{Repair Insufficiency:} In rare cases, even the full repair loop cannot restore 
satisfiability without retracting many commitments. When $\Delta_{\text{past}} > 3$, we 
abandon repair and mark the entire bundle as PARTIAL (partial success), reporting reduced 
\textsc{SetConsRate} for that case.

\end{document}